\documentclass[10pt,conference,final,comsoc]{IEEEtran}
\IEEEoverridecommandlockouts
\usepackage{cite}
\usepackage{amsmath,amssymb,amsfonts}
\usepackage{algorithmic}
\usepackage{algorithm}
\usepackage{graphicx}
\usepackage{textcomp}

\usepackage{xcolor}
\usepackage[letterpaper, top=0.75in, bottom=1.05in, left=0.625in, right=0.625in]{geometry}
\def\BibTeX{{\rm B\kern-.05em{\sc i\kern-.025em b}\kern-.08em
    T\kern-.1667em\lower.7ex\hbox{E}\kern-.125emX}}
\begin{document}

\title{Mixed-Precision Federated Learning via Multi-Precision Over-the-Air Aggregation

\thanks{\textsuperscript{*}Corresponding author. The work is supported by EPSRC CHEDDAR: Communications Hub for Empowering Distributed clouD computing Applications and Research (EP/X040518/1) (EP/Y037421/1). We acknowledge Dr. Yun Wu at QUB for his inspiration.}
}

\author{\IEEEauthorblockN{Jinsheng Yuan*}
\IEEEauthorblockA{\textit{Faculty of Engineering \& Applied Sciences} \\
\textit{Cranfield University}\\
United Kingdom \\
jinsheng.yuan@cranfield.ac.uk}
\and
\IEEEauthorblockN{Zhuangkun Wei}
\IEEEauthorblockA{\textit{Department of Engineering} \\
\textit{Durham University}\\
United Kingdom \\
zhuangkun.wei@durham.ac.uk}
\and
\IEEEauthorblockN{Weisi Guo}
\IEEEauthorblockA{\textit{Faculty of Engineering \& Applied Sciences} \\
\textit{Cranfield University}\\
United Kingdom \\
weisi.guo@cranfield.ac.uk}
}

\maketitle

\begin{abstract}
Over-the-Air Federated Learning (OTA-FL) is a privacy-preserving distributed learning mechanism, by aggregating updates in the electromagnetic channel rather than at the server. A critical research gap in existing OTA-FL research is the assumption of homogeneous client computational bit precision. While in real world application, clients with varying hardware resources may exploit approximate computing (AxC) to operate at different bit precisions optimized for energy and computational efficiency. Model updates with varying precisions among clients present a significant challenge for OTA-FL, as they are incompatible with the wireless modulation superposition process. Here, we propose an mixed-precision OTA-FL framework of clients with multiple bit precisions, demonstrating the following innovations: (i) the superior trade-off for both server and clients within the constraints of varying edge computing capabilities, energy efficiency, and learning accuracy requirements compared to homogeneous client bit precision, and (ii) a multi-precision gradient modulation scheme to ensure compatibility with OTA aggregation and eliminate the overheads of precision conversion. Through case study with real world data, we validate our modulation scheme that enables AxC based mixed-precision OTA-FL. In comparison to homogeneous standard precision of 32-bit and 16-bit, our framework presents more than 10\% in 4-bit ultra low precision client performance and over 65\%and 13\% of energy savings respectively. This demonstrates the great potential of our mixed-precision OTA-FL approach in heterogeneous edge computing environments.
\end{abstract}


\section{Introduction}

\begin{figure}[htbp]
\centering
\includegraphics[width=\linewidth]{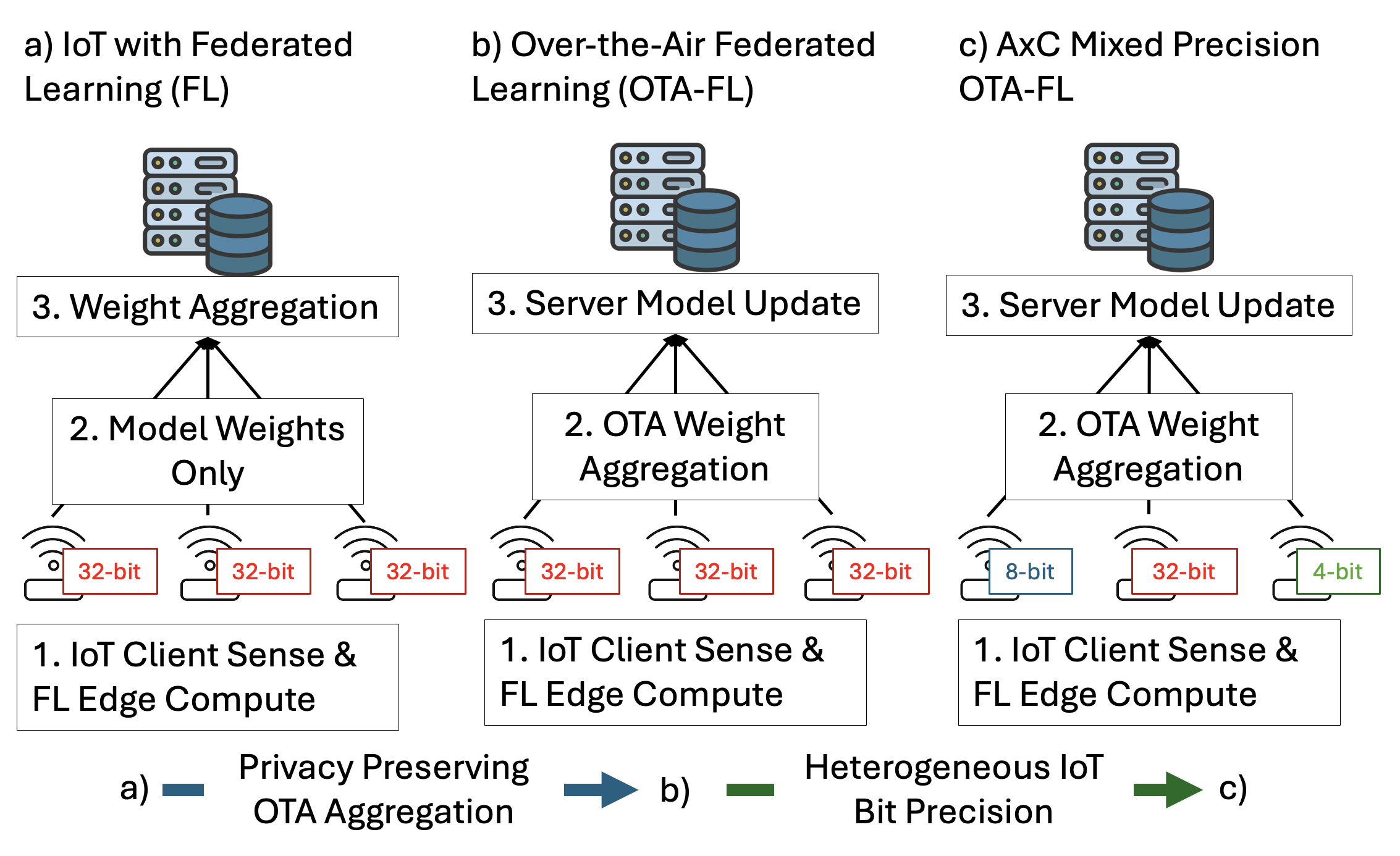}
\caption{End-to-end Federated Learning (FL) system moves from (a) FL, to (b) privacy preserving OTA-FL, to (c) energy efficient AxC OTA-FL. The research challenge from (b) to (c) is to achieve heterogeneous OTA weight aggregation to cater for mixed bit precision IoT edge computation.}
\label{fig_1}
\end{figure}

Federated Learning (FL) \cite{McMahan2016FL}, see in Fig.~\ref{fig_1}a, has emerged as a widely studied and applied distributed learning framework that ensures security and privacy by sharing and aggregating model parameters instead of raw data, as is done in centralized learning. Recent advancements in FL research \cite{Chen2024Advancements} primarily focus on two key aspects: (i) enhancing privacy and security, and (ii) improving efficiency.

Over-the-Air Federated Learning (OTA-FL) \cite{Yang2020OTAFL}, see in Fig.~\ref{fig_1}b, represents a novel paradigm in FL dedicated for wireless networks. By leveraging the inherent randomness of physical-layer channel states and electromagnetic superposition for aggregating model updates, OTA-FL enhances privacy preservation without the additional computational overhead associated with other approaches, such as those employing differential privacy mechanisms \cite{el2022differential}. Recent advancements in OTA-FL encompass transmitting node precoding\cite{sery2021over}, server beamforming vector optimization\cite{kim2023beamforming}, and reconfigurable intelligent surface (RIS) phase adjustment\cite{zheng2022balancing}.

\begin{figure*}[!t]
\centering
\includegraphics[width=1\textwidth]{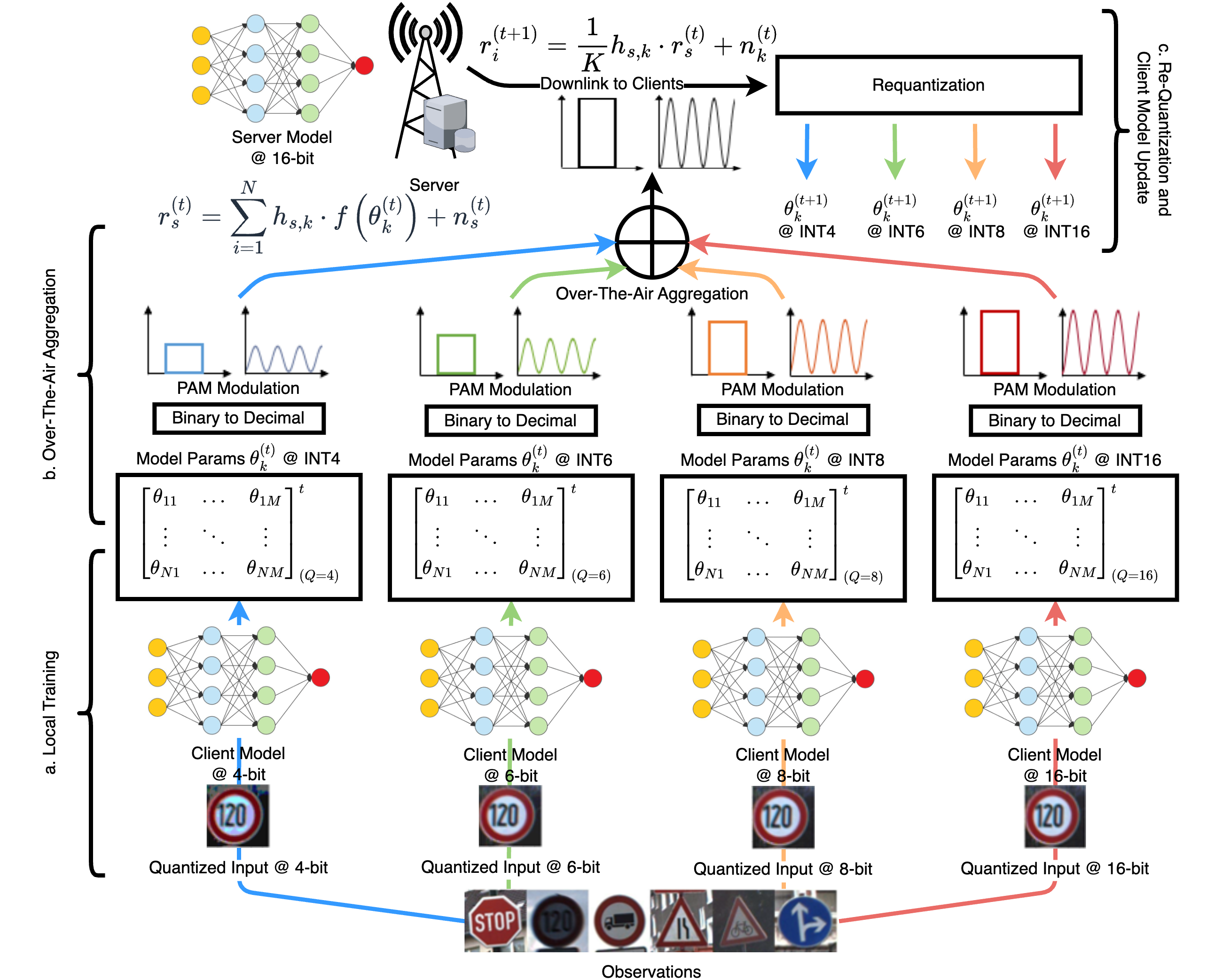}
\caption{Structure of our proposed Approximate Computing (AxC) based OTA-FL framework of multi-precision clients and unified multi-precision modulation scheme. The intelligent transport validation case study here is a multi-precision federated supervised traffic sign recognizer. (a) Clients operate end-to-end at their designated computation precisions with their own data and labels. (b) Multi-precision OTA aggregation process (uplink). (c) Downlink, re-quantization and client model update.}
\label{fig_sys_dgm}
\end{figure*}

\textbf{Related Works:} In Federated Learning (FL) systems, optimizing the balance between task performance and resource constraints such as communication, computation, and energy has been a key area of focus. Strategies aimed at reducing communication overhead include SCAFFOLD \cite{karimireddy2020scaffold}, which uses control variates to correct for client drift, and LoSAC \cite{chen2023losac}, which locally updates the estimate for the global full gradient after each local model update to enhance communication efficiency. These methods achieve faster convergence with fewer communication rounds. On the computation side, various approaches have been explored, including hardware acceleration \cite{Zhang2015Optimizing}, network architecture optimization \cite{Elsken2019NAS}, model slicing that assigns sub-models to clients based on their hardware capabilities \cite{lee2024recurrent}, and Approximate Computing (AxC) methods.

\textbf{Inspiration and Motivation:} Approximate Computing (AxC) methods encompass both hardware and software domains, aiming to balance task performance with energy efficiency and computational resource optimization. A significant source of inspiration for us within the AxC domain arises from Field Programmable Gate Arrays (FPGA) accelerators \cite{Flegar2019FloatX}. Unlike traditional Central Processing Units (CPUs) and Graphics Processing Units (GPUs), which operate at fixed preset precision levels, FPGAs provide a customizable computation paradigm. This flexibility of reprogrammability allows for highly efficient operations at varying precision levels across diverse applications, including networking, signal processing, multimedia codec operations, and machine learning acceleration \cite{mittal2016survey}. In this work, we investigate the potential of heterogeneous quantization to address computational and communication trade-offs in Federated Learning (FL). The research focuses on two key questions: (i) what are the potential gains in computation, energy efficiency, and performance when using heterogeneous client precision levels compared to conventional homogeneous approaches, and (ii) how to design an effective modulation scheme that supports multi-precision client updates in Over-the-Air (OTA) aggregation.

\textbf{Contribution:} In this paper, our contributions are threefold:
\begin{enumerate}
    \item We propose an AxC-based OTA-FL framework for multi-precision clients, aimed at bolstering performance and efficiency - see Fig.~\ref{fig_1}(c).
    \item We develop heterogeneous gradient resolution modulation schemes to ensure compatibility with physical-layer OTA aggregation and eliminate the overheads of precision conversion - see Fig.~\ref{fig_sys_dgm}(b).
    \item We conducted a case study with real-world data to demonstrate the effectiveness of our approach in comparison to homogeneous precision OTA-FL systems. Results show a notable improvement in server convergence speed and more than 10\% in 4-bit ultra low precision client performance and over 65\% and 13\% of energy savings respectively compared to FL with homogeneous 32-bit and 16-bit clients. These findings highlight the potential of our framework in resource-constrained, heterogeneous computing environments.
\end{enumerate}

\textbf{Paper Structure:} The remainder of the paper is organized as follows: Section II provides a detailed exploration of the system setup of our mixed-precision OTA-FL. Section III delves into the design specifics of our AxC-based OTA-FL framework. Experiments and results are presented in Section IV. Section V concludes the paper, summarizing key findings and contributions.

\section{System Setup}

\subsection{Federated Learning}

We consider an over-the-air federated learning system with $N$ clients, at each communication round, $K$ of them are selected to update denoted by $\mathcal{K} = \{1, ..., K\}$. Each client $k$ has a local dataset $\mathcal{D}_k$. For $\mathcal{T}$ communication rounds, the clients collaboratively refine a unified global model, while preserving the privacy of their local data. This process collects the local trained models of clients, and aggregates these models to update the global model. The refined global model is then distributed back to the edge devices for subsequent predictions and further local training iterations, thereby progressively enhancing the global model's accuracy. The model aggregation phase, central to FL, can be mathematically represented as:

\begin{equation}
\theta^{(t+1)}=\frac{1}{K}\sum_{k=1}^K w_k \cdot \theta_k^{(t)},
\label{eq_fedavg}
\end{equation}

where $\theta^{(t+1)}$ denotes the parameters of the global model after the $(t+1)$-th training iteration, $\theta_k^{(t)}$ represents the parameters of the local model from the edge device $k$ at the $t$-th communication round, and $w_k$ signifies the relative contribution (or weight) of the edge device $k$, typically proportional to its dataset size.

\subsection{Over-The-Air (OTA) Computation for FL}

The principle of Over-The-Air (OTA) computation\cite{nazer2007computation}, exploits the natural superposition property inherent to wireless channels. Within the FL framework, this concept finds practical application during the gradient aggregation phase. By utilizing a common uplink bandwidth across all edge devices for gradient transmission, the superposition property of the channel facilitates direct aggregation. Consider a Single-Input Single-Output (SISO) fading channel between the server and an edge device $k$, characterized by a Rayleigh distributed random variable $h_{s,k}\in\mathbb{C}$. The OTA aggregation process can then be modeled as:

\begin{equation}
r_s^{(t)}=\sum_{k=1}^K h_{s,k}\cdot f\left(\theta_{k}^{(t)}\right)+n_s^{(t)},
\end{equation}
where $f(\cdot)$ embodies the comprehensive process including source coding, constellation design, up-sampling, modulation, and precoding. Here, $r_s^{(t)}$ represents the aggregated signal received by the server in the $t$-th upload cycle, and $n_s^{(t)}$ denotes the additive noise.

In contrast to traditional federated learning aggregation mechanisms, leveraging OTA computation necessitates careful design of $f(\cdot)$ to address two critical challenges: (i) mitigating the effects of channel fading, and (ii) ensuring accurate linear summation of gradients from multiple edges. This becomes particularly complex when considering edges with disparate computation, storage, and operation precision constraints. A notable challenge arises in aggregating gradients quantized at different levels, exemplified by the non-commutative property of quantized modulations:
\begin{equation}
QAM([\theta_j]_{q_j}) + QAM([\theta_k]_{q_k}) \neq QAM([\theta_j]_{q_j} + [\theta_k]_{q_k}),
\end{equation}
where $QAM(\cdot)$ represent the quadrature modulation function and $[\theta_k]_{q_k}$ indicates model parameters of client $k$ is at $q_k$ quantization levels (in terms of bits) utilized for encoding weights and biases of models at different edge devices.

\subsection{Approximate Computing and Low-Precision ML} 
Within AxC methods, quantization stands out for its universal applicability, and compatibility with existing FL systems. In the context of edge computing, FPGA accelerators become particularly advantageous, as they can be dynamically reprogrammed to efficiently perform computation at any custom precision level including 4-bit and below, thus significantly reducing computation demand and energy consumption in resource-constrained environments \cite{Zhang2015Optimizing}. 

\begin{table}[ht]
    \centering
    \caption{GTSRB Classification Performance of Common Models across Quantization Levels}
    \label{tab_benchmark}
    \begin{tabular}{lccccc}
    \hline
    Model & 8-bit & 6-bit & 4-bit & 3-bit & 2-bit \\
    \hline
    densenet\_161 & 96.56\% & 96.45\% & 91.55\% & \textbf{\textcolor{red}{39.83\%}} & \textbf{\textcolor{red}{0.00\%}} \\
    efficient\_net\_b4 & 94.77\% & 94.74\% & 90.55\% & \textbf{\textcolor{red}{42.24\%}} & \textbf{\textcolor{red}{0.07\%}} \\
    efficient\_net\_v2m & 95.26\% & 95.19\% & 85.75\% & \textbf{\textcolor{red}{7.68\%}} & \textbf{\textcolor{red}{6.22\%}} \\
    reg\_net\_x\_16gf & 96.56\% & 96.11\% & \textbf{\textcolor{orange}{80.84\%}} & \textbf{\textcolor{red}{0.00\%}} & \textbf{\textcolor{red}{0.00\%}} \\
    reg\_net\_y\_3\_2gf & 93.53\% & 92.75\% & \textbf{\textcolor{orange}{72.72\%}} & \textbf{\textcolor{red}{6.97\%}} & \textbf{\textcolor{red}{0.00\%}} \\
    resnet\_50 & 94.94\% & 94.33\% & \textbf{\textcolor{orange}{65.21\%}} & \textbf{\textcolor{red}{0.00\%}} & \textbf{\textcolor{red}{0.83\%}} \\
    squeeze\_net\_1\_0 & 87.29\% & 85.41\% & \textbf{\textcolor{orange}{72.95\%}} & \textbf{\textcolor{red}{39.85\%}} & \textbf{\textcolor{red}{6.64\%}} \\
    \hline
    \multicolumn{6}{l}{\textbf{\textcolor{orange}{Orange:}} damaged but usable performance ($65-85\%$)}\\
    \multicolumn{6}{l}{\textbf{\textcolor{red}{Red:}} unacceptable performance ($<65\%$).}\\
    \end{tabular}
\end{table}

As quantization can significantly reduce the resource demand for storing, inferring, and training machine learning models \cite{gupta2015deep, Hubara2016BinaryNet} at the price of performance degradation, it is essential to demonstrate the trade-offs of low-precision machine learning. We benchmarked the performance degradation of common CNN models in quantization, as shown in Table.~\ref{tab_benchmark}. All these models are trained in 32-bit floating-point format and then quantize to lower bit-levels. The degradation only becomes noticeable when quantized to 8-bit, and retains acceptable before being quantized below 4-bit. 

It is also worth noting that such performance can not be stably achieved via conventional end-to-end training at the same lowest precision level due to the limit of gradient dynamic range and cumulative error of low precision, and additionally, when the size of dataset increases, gradients may require a larger dynamic range to fit the data. 

Although end-to-end low precision training can be achieved through dedicated design of format, quantization algorithms, and arithmetic implementation \cite{Sun2020ULPDNN}, a key advantage of our mixed-precision OTA-FL framework is simplicity for application. We leverage mixed-precision federated learning, to enable the low precision ML on edge devices and hence achieve superior trade-off of computation and performance, and reduction of energy consumption.

\section{Mixed-Precision OTA Federated Learning}

The process of our Mixed-Precision OTA Federated Learning framework can be described as Algorithm~\ref{alg:framework}.

\begin{algorithm}
\caption{Mixed-Precision OTA Federated Learning}
\label{alg:framework}
\begin{algorithmic}[1]
    \STATE \textbf{Input:} Initial global model $\theta^{(0)}$, number of communication rounds $T$, number of clients $K$
    \STATE \textbf{Output:} Trained global model $\theta^{(T)}$
    
    \FOR {each round $t = 1$ to $T$}
        \STATE \textbf{Step 1: Broadcast global model $\theta^{(t-1)}$}
        \STATE Server broadcasts global model $\theta^{(t-1)}$ to all clients

        \STATE \textbf{Step 2: Local training at clients}
        \FOR {each client $k = 1$ to $K$ (in parallel)}
            \STATE Quantize $\theta^{(t-1)}$ to designated precision level $q_k$
            \STATE Local training $[\theta_k^{(t)}]_{q_k} = \text{Training}([\theta^{(t-1)}]_{q_k}, \mathcal{D}_k)$
            \STATE Calculate update $\Delta [\theta_k^{(t)}]_{q_k} = [\theta_k^{(t)}]_{q_k} - [\theta^{(t-1)}]_{q_k}$
        \ENDFOR

        \STATE \textbf{Step 3: Over-the-Air aggregation of mixed-precision model updates}
        \FOR {each client $k = 1$ to $K$ (in parallel)}
            \STATE Convert model update $\Delta [\theta_k^{(t)}]_{q_k}$ to decimal
            \STATE Amplitude modulation, channel estimation and uplink 
        \ENDFOR

        \STATE \textbf{Step 4: Server-side post-processing and downlink}
        \STATE Received signal $r_s^{(t)}\approx\sum_{k=1}^K \theta_k^{(t)}$
        \STATE Broadcast updated model $\frac{r_s^{(t)}}{K}\approx\frac{\sum_{k=1}^K \theta_k^{(t)}}{K}$
    \ENDFOR

    \STATE \textbf{Return} final global model $\theta^{(T)}$
\end{algorithmic}
\end{algorithm}

\subsection{Multi-Precision Over-The-Air Aggregation} 

Over-The-Air Federated Learning (OTA-FL) encompasses a three-step process during each update round, denoted as the $t$-th round. The sequence begins with each client $k$ conducting local training and producing a model update $\Delta [\theta_k^{(t)}]_{q_k}$. Subsequently, as shown in Fig.~\ref{fig_sys_dgm} (b), the client prepares for transmission by converting binary parameters of its designated precision into decimal equivalents. These decimal values are then modulated onto carrier waves through amplitude modulation, creating signals ready for bandwidth transmission. Our modulation function can be described as follows.

\begin{equation}
M([\theta_k]_{q_k}) = [\theta_k]_{q_k} \cdot cos2\pi f_c t
\end{equation}

where $M(\cdot)$ represent the modulation function, $[\theta_i]_{q_i}$ indicates model parameters of client $k$ is at $q_k$ quantization levels, and $f_c$ is the channel frequency.

In the upload phase, the client $k$ performs the channel estimation to determine the communication link, denoted as $h_{s,k}$, between the server and itself. This estimated channel information is crucial for implementing transmission beamforming, a technique employed for channel compensation. Such compensation is integral to facilitating efficient OTA aggregation of transmitted data. The procedural intricacies of this approach are elucidated below.

\subsubsection{Channel Estimation at Clients}
Channel estimation between the server and each client is done by broadcasting a predefined pilot sequences $u$ from the server. Then, client $k$ can estimate the channel between $k$ and the server $h_{s,k}$ as follows.

\begin{equation}
\hat{h}_{s,k}=y_s\cdot \frac{u^*}{|u|^2}\approx h_{s,k},
\end{equation}

where $y_s=h_{s,k}\cdot u+n_s$ is the received signals at server with receiving noise $n_s$.  

\subsubsection{Uplink Design}
After channel estimation at each client $k$, the clients modulate model updates to carrier frequency and compensate for channel distortion via its estimated channel. The base-band transmitted signal is designed as follows.

\begin{equation}
f\left(\theta_k^{(t)}\right)=\hat{h}_{s,k}^{-1}\cdot \theta_k^{(t)}.
\end{equation}

Hence, aggregation can be done via the natural superposition of the electromagnetic wave, i.e., $r_s^{(t)}\approx\sum_{k=1}^K \theta_k^{(t)}$. 

\subsubsection{Downlink Design}
After OTA aggregation, the server broadcasts the updated model, i.e., $r_s^{(t)}/K\approx\sum_{k=1}^K \theta_k^{(t)}/K$ back to clients. The received signal at each client $k$ is:

\begin{equation}
r_k^{(t+1)}= \frac{1}{K} h_{s,k} \cdot r_s^{(t)}+n_k^{(t)}. 
\end{equation}

The client $i$ then recovers the aggregated gradient via the estimated channel, as follows. 

\begin{equation}
\tilde{\theta}_k^{(t+1)}=h_{s,k}^{-1} \cdot r_k^{(t+1)}\approx \frac{1}{K} \sum_{k=1}^K \theta_k^{(t)},
\end{equation}

where $\tilde{\theta}_k^{(t+1)}$ is the updated gradient of client $k$, and will be local trained for the $(t+1)$th round. 

Furthermore, during the OTA aggregation process, the mixed-precision quantization scheme, which remains inherently opaque to potential adversaries, significantly enhances the system’s security against attacks targeting the aggregation process. 

\subsection{Approximate Computing via Quantization} 

\begin{algorithm}
\caption{Quantization Function}
\label{alg:quantization}
\begin{algorithmic}[1]
    \STATE \textbf{Input:} Tensor $W \in \mathbb{R}^{m \times n}$, type: "fixed-point" or "floating-point", bit-width $b$
    \STATE \textbf{Output:} Quantized tensor $Q \in \mathbb{R}^{m \times n}$
    \IF {type is "fixed"}
        \STATE $w_{\text{min}} = \min(W)$, $w_{\text{max}} = \max(W)$
        \STATE $\text{scale} = \frac{w_{\text{max}} - w_{\text{min}}}{2^b - 1}$, $\text{zero\_point} = -\frac{w_{\text{min}}}{\text{scale}}$
        \FOR {each $w_{ij}$ in $W$}
            \STATE $q_{ij} = \max(0, \min(2^b - 1, \left\lfloor \frac{w_{ij}}{\text{scale}} + \text{zero\_point} \right\rfloor))$
        \ENDFOR
    \ELSIF {type is "floating-point"}
        \FOR {each $w_{ij}$ in $W$}
            \STATE Truncate mantissa and exponent to fit $b$ bits
        \ENDFOR
    \ENDIF
    \STATE \textbf{Return} $Q$
\end{algorithmic}
\end{algorithm}

To ensure broad applicability, we employ a simple and efficient quantization algorithm, as outlined in Algorithm~\ref{alg:quantization}. For precision levels of 8-bit and higher, both fixed-point and floating-point formats are supported. However, fixed-point format is preferred for lower precision levels due to the limited dynamic range of floating-point formats under 8-bit representation.

On the client side, the quantization function is systematically applied to every layer of the CNN model, spanning from input to output, and is integrated into both the forward and backward passes. This approach ensures a unified precision level throughout the end-to-end system.

\subsection{Energy Consumption Estimation}

Along with compute savings, as an equally important product of quantization, we also measure the energy savings which originate from the higher throughput of operations at lower precisions. We estimate the energy consumption of training a ResNet-50 model at multiple precision levels between 32-bit and 4-bit on 9 Xilinx FPGA platforms of varying specifications. Then we use the average relative energy saving compared to 32-bit at each precision levels of all these platforms to estimate the total savings for our mixed-precision clients.

However, the energy consumption for the same task of FPGA platforms can vary significantly based on the program design of the same hardware. In FPGA focused researches, accurate energy consumption can be obtained through the official power analysis toolkit \cite{xilinx_xpe_2024}. For this paper, we provide a modest energy consumption estimation for the clients based on official data sheets \cite{amd_ultrascale_2021} of typical FPGA edge platforms with Equation~\ref{eq_gops} below to showcase the potential of energy savings of our approach. 

\begin{equation}
    E_{ML}=\frac{D_{ML}}{F_{DSP}\cdot N_{DSP}\cdot N_{MAC}}\cdot E_{Package}
    \label{eq_gops}
\end{equation}

Where $E_{ML}$ and $D_{ML}$ represent the energy consumption and computation demand by operations per communication round of ML task respectively, $F_{DSP}$ is the frequency of DSP slices, $N_{DSP}$ is the number of onboard DSP slices, and $N_{MAC}$ is the number of multiply-accumulate (MAC) operations each DSP slice can carry out per cycle, and $E_{Package}$ represent typical package energy consumption\cite{abdelhamid2023quantitative}.

\section{Experiments}

This section details the experiment setup and results addressing the research questions posited earlier, focusing on (i) the efficacy of multi-precision OTA aggregation, and (ii) the formulation of client quantization schemes to optimize performance across both server and client dimensions. 

\subsection{Settings}

Our experimental framework emulates an OTA-FL system of 15 clients working at designated quantization levels. The model structure is ResNet-50 with ImageNet \cite{deng2009imagenet} pre-trained weights initialization. The FL operates over 100 communication rounds with 5-30dB of emulated Gaussian noise.

\subsubsection{Data}

We utilize the German Traffic Sign Recognition Benchmark (GTSRB) \cite{Stallkamp2012GTSRB} as the dataset for our experiments. This dataset consists of 39,209 training samples and 12,630 testing samples across 43 traffic sign classes, captured in real-world driving environments. It reflects the variability of real-world conditions, including changes in lighting, weather, perspectives, and occlusions. This diversity ensures the dataset’s authenticity and makes it highly suitable for advancing research and development in smart transportation systems. In our experimental setup, each client is assigned an equal subset of the data. 

\subsubsection{Quantization Schemes}

We assign quantization levels to the 15 clients by a group of 5. Each scheme consists of 3 precision levels, and each precision level is assigned to 5 clients. Quantization levels are chosen from $[32, 24, 16, 12, 8, 6, 4]$.

\subsubsection{Performance Metrics}

Evaluation metrics for the server include convergence speed, measured by the number of communication rounds the system took to converge, and final performance of the aggregated model. For clients, we assess their performances after aggregation and re-quantization. 

\subsubsection{Energy Efficiency}

We compare the energy savings of our mix-precision clients with homogeneous precision clients at standard 32-bit, 16-bit, 8-bit and 4-bit. 

\subsection{Results}

\subsubsection{Energy Efficiency Estimation}

\begin{table}[h!]
\centering
\caption{Estimated Energy Consumption per Sample for ResNet-50 Forward Pass and Relative Savings Compared to 32-bit}
\label{tab:energy_est}
\begin{tabular}{lcccccc}
\hline
 & 32-bit & 16-bit & 12-bit & 8-bit & 6-bit & 4-bit \\ \hline
Energy Cost (J) & 0.36 & 0.17 & 0.16 & 0.022 & 0.021 & 0.0056 \\ \hline
Saving (\%) & 0 & 52.58 & 56.15 & 93.89 & 94.17 & 98.45 \\ \hline
\end{tabular}
\end{table}

Based on Eq.~\ref{eq_gops}, we estimate energy consumption of forward passing one sample through a ResNet-50 network, across following quantization levels $[32, 24, 16, 12, 8, 6, 4]$ on 9 platforms of different hardware resources such as logic cells and DSP slices. In Table~\ref{tab:energy_est}, we present the average energy cost and relative savings to 32-bit of these platforms. Notably, due to under-utilization of hardware, quantizing to 16-bit and 12-bit share very similar degree energy saving, and the same applies to 8-bit and 6-bit. From the table, we can also see the diminishing energy saving gain when further quantizing from low precision like 8-bit to ultra low ones like 4-bit.

\subsubsection{Federated Training and Server Performance}

\begin{figure}[htbp]
\centering
\includegraphics[width=1\linewidth]{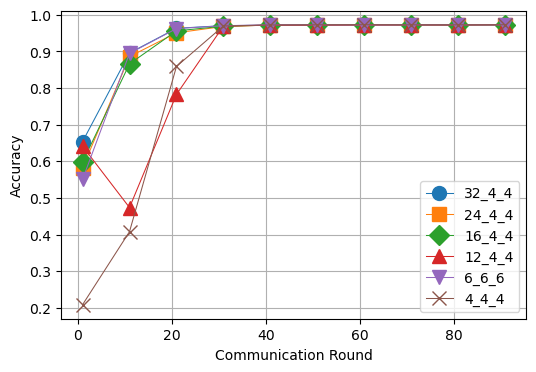}
\caption{Training accuracy in 100 communication rounds, with ImageNet pre-trained weights initialization, scheme $[4, 4, 4]$ (denoted by brown 'X'), and scheme $[12, 4, 4]$ (denoted by red upright triangle) converge significantly slower than other schemes, even when the latter one has a better random start.}
\label{fig_lowest_qlvl_train_acc}
\end{figure}

The convergence velocity, as depicted in Fig.~\ref{fig_lowest_qlvl_train_acc}, indicates that setups of uniform 4-bit clients, or a mixed-precision schema of $[12, 4, 4]$ bits, exhibit slower and more erratic initial convergence, even when the latter has a better random start in training. In contrast, setups incorporating clients with 16-bit precision or higher demonstrate more rapid and stable convergence, achieving approximately 90\% accuracy within 10 communication rounds. Notably, for clients of high resource capacity, 32-bit or 24-bit precision only offers marginal training gains compared to 16-bit precision. The server model performance of all quantization schemes reached 97\% top-1 accuracy within a tight 0.3\% margin after 100 communication rounds, underscoring the effectiveness of the federated learning framework in achieving high accuracy with mixed-precision clients.

\subsubsection{Client Performance}

\begin{figure}[htbp]
\centering
\includegraphics[width=1\linewidth]{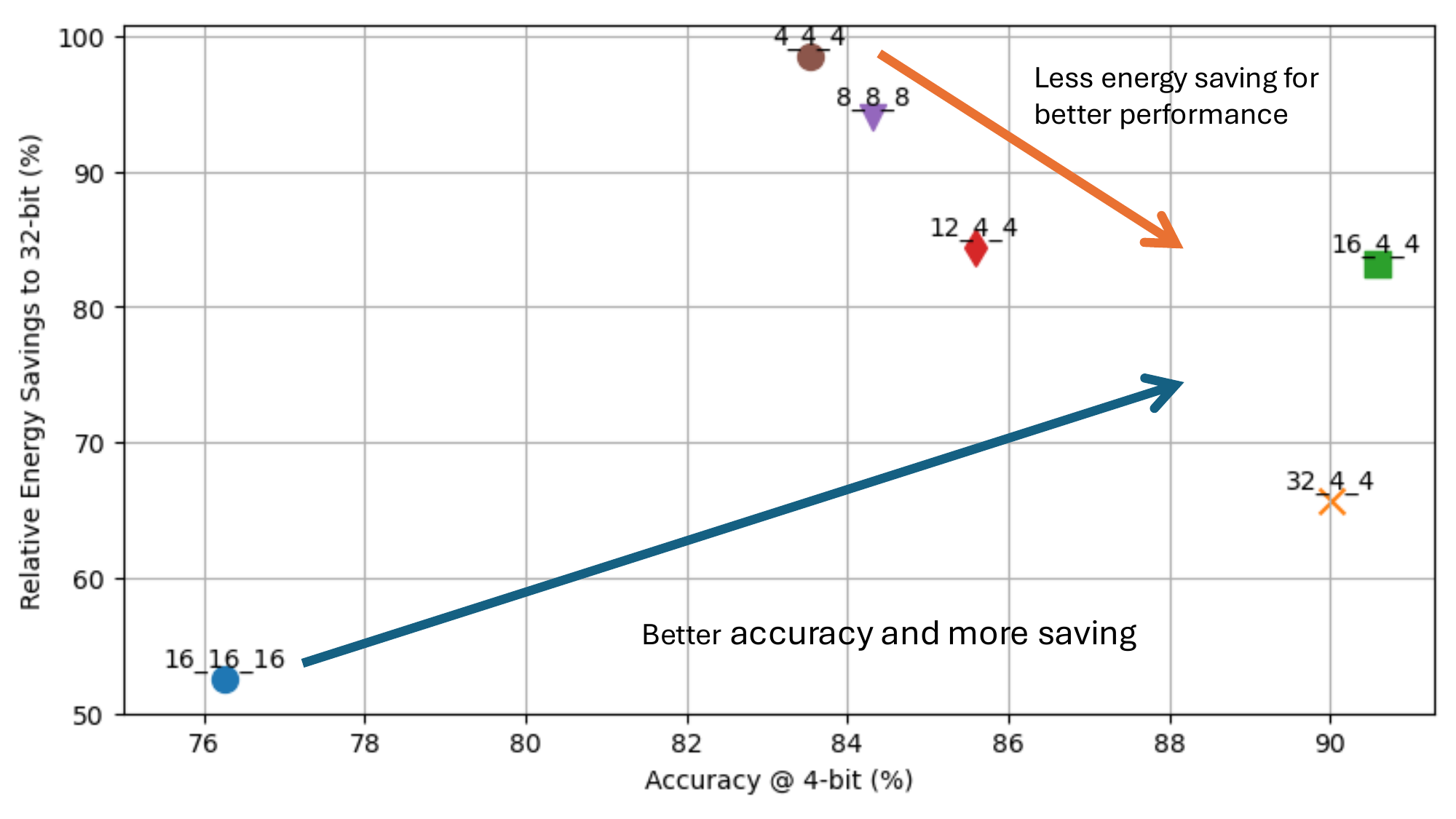}
\caption{Trade-offs between accuracy of model quantized to 4-bit and energy savings in comparison to homogeneous 32-bit and 16-bit clients, schemes near bottom right corner presents superior trade-off towards accuracy.}
\label{fig_tradeoff}
\end{figure}

After 100 rounds, the final global model, $\theta^{100}$, is broadcast to clients. Here, we focus on the clients at the lowest precision, 4-bit, for higher counterparts have better performance and minor degradation from well converged global model as illustrated in Table~\ref{tab_benchmark} and Fig.~\ref{fig_lowest_qlvl_train_acc}. 

As shown in Fig.~\ref{fig_tradeoff}, in comparison to FL systems of homogeneous clients at 32-bit and 16-bit, based on our estimation in Table~\ref{tab:energy_est}, our FL with mixed-precision clients models can save over 65\% and 13\% of energy consumption respectively, while gaining more than 10\% in accuracy on clients at 4-bit. Notably, for those under schemes incorporating 16-bit precision or higher, when re-quantized for 4-bit clients, attain around 5\% higher accuracy, and this performance boost for lower precision clients from higher precision counterparts shows diminishing returns beyond 16-bit precision. While comparing to FL of homogeneous clients at 8-bit and 4-bit, our mixed-precision FL can trade mere 10\% of energy savings for 5\% of accuracy.

\section{Conclusion}
In this study, we introduce a framework for Over-the-Air Federated Learning (OTA-FL) that incorporates Approximate Computing (AxC) to accommodate clients of multiple precision levels. Our novel mixed-precision OTA aggregation mechanism enables the enhancement of overall performance, computational and energy efficiency within federated learning systems, especially for those of ultra low precision. Our framework is universally applicable and compatible with existing FL systems, unveiling the huge energy saving potential of incorporating clients at ultra low precision while having their performance improved. These advantages of leveraging multi-precision client configurations in OTA-FL systems, in both performance and energy savings, are particularly valuable in resource-diverse and heterogeneous edge computing environments. This study may serve as a foundational guideline for the architectural design and optimization of future green and sustainable multi-precision OTA-FL systems, at the cost of a less efficient mixed-precision modulation scheme for OTA aggregation. In addition to FL systems, the mixed-precision modulation scheme can also facilitate other distributed computation applications tailored for wireless networks.

\bibliographystyle{IEEEtran}
\bibliography{IEEEabrv, reference}

\end{document}